\pdfoutput=1

\documentclass[11pt]{article}

\usepackage{acl}

\usepackage{times}
\usepackage{latexsym}

\usepackage[T1]{fontenc}

\usepackage[utf8]{inputenc}

\usepackage{microtype}

\usepackage{graphicx}
\usepackage{ulem}
\usepackage{mathrsfs}
\usepackage{amsmath}
\usepackage{booktabs}
\usepackage{multirow}
\usepackage{amsfonts}

\title{A Double-Graph Based Framework for Frame Semantic Parsing}

\author{Ce Zheng$^{1}$, Xudong Chen$^{1,2}$, Runxin Xu$^{1}$, Baobao Chang$^{1}$\thanks{\;\;Corresponding author} \\
  $^{1}$The MOE Key Laboratory of Computational Linguistics, Peking University, China \\
  $^{2}$School of Software and Microelectronics, Peking University, China \\
  \texttt{\{zce1112zslx,xdc\}@pku.edu.cn} \\\texttt{runxinxu@gmail.com; chbb@pku.edu.cn } \\
  }

\begin{document}
\maketitle
\begin{abstract}
Frame semantic parsing is a fundamental NLP task, which consists of three subtasks: frame identification, argument identification and role classification. Most previous studies tend to neglect relations between different subtasks and arguments and pay little attention to ontological frame knowledge defined in FrameNet. In this paper, we propose a \uline{K}nowledge-guided \uline{I}ncremental semantic parser with \uline{D}ouble-graph (\textsc{Kid}). We first introduce Frame Knowledge Graph (FKG), a heterogeneous graph containing both frames and FEs (Frame Elements) built on the frame knowledge so that we can derive knowledge-enhanced representations for frames and FEs. Besides, we propose Frame Semantic Graph (FSG) to represent frame semantic structures extracted from the text with graph structures. In this way, we can transform frame semantic parsing into an incremental graph construction problem to strengthen interactions between subtasks and relations between arguments. Our experiments show that \textsc{Kid} outperforms the previous state-of-the-art method by up to 1.7 F1-score on two FrameNet datasets. Our code is availavle at \url{https://github.com/PKUnlp-icler/KID}.

\end{abstract}

\section{Introduction}

\begin{figure}[t]
\centering
\includegraphics[width=0.9\columnwidth]{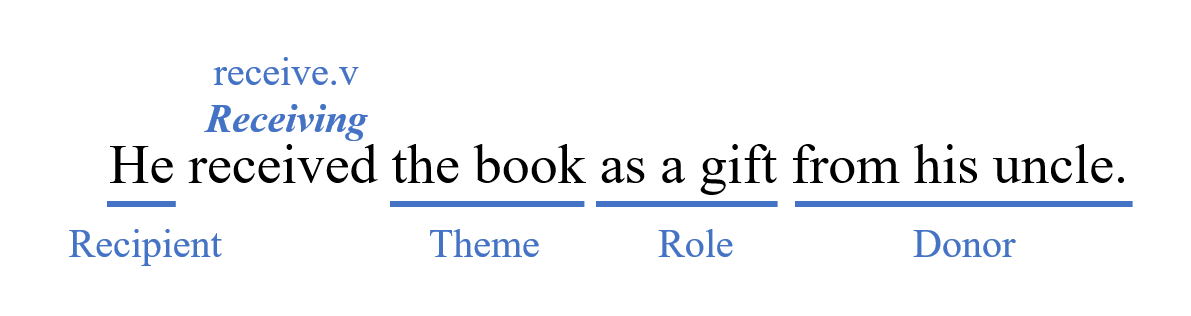}
\caption{Given the target \textbf{receive} in this sentence, the frame identification is to identify the frame \textbf{\textit{Receiving}} evoked by it; the argument identification is to find the arguments (\textit{He}, \textit{the book}, ...) of this target; the role classification is to assign frame elements (\textbf{Recipient}, \textbf{Theme}, ...) as semantic roles to these arguments.}
\label{fig1}
\end{figure}

The frame semantic parsing task \citep{gildea2002automatic, baker-etal-2007-semeval} aims to extract frame semantic structures from sentences based on the lexical resource FrameNet \citep{baker-etal-1998-berkeley-framenet}. As shown in Figure \ref{fig1}, given a target in the sentence, frame semantic parsing consists of three subtasks: frame identification, argument identification and role classification. Frame semantic parsing can also contribute to downstream NLP tasks such as machine reading comprehension \citep{guo-etal-2020-frame}, relation extraction \citep{zhao2020cfsre} and dialogue generation \citep{gupta-etal-2021-controlling}.

\begin{figure*}[t]
\centering
\includegraphics[width=1.7\columnwidth]{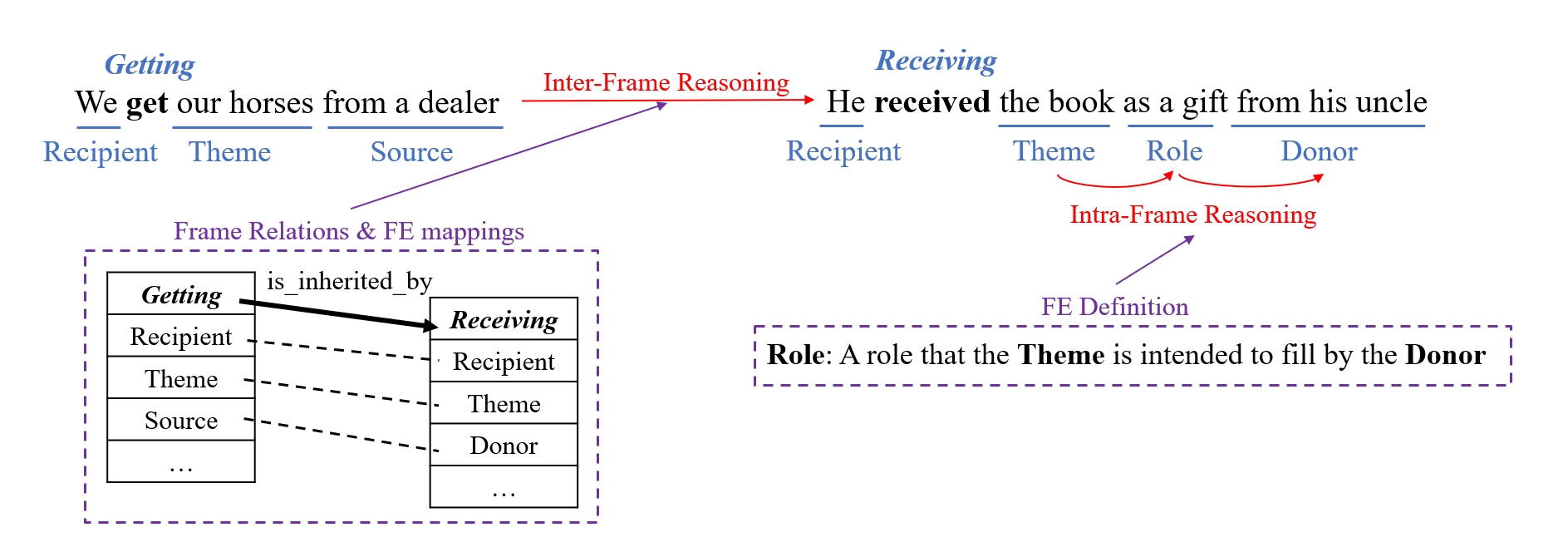}
\caption{An example of how frame knowledge contributes to frame semantic parsing. The frame semantic relations and FE mappings guide inter-frame reasoning (from left sentence to right); and the FE definitions help with intra-frame reasoning (\textbf{Theme} to \textbf{Role} and \textbf{Role} to \textbf{Donor}).}
\label{fig2}
\end{figure*}

FrameNet is an English lexical database, which defines more than one thousand hierarchically-related frames to represent situations, objects or events, and nearly 10 thousand FEs (Frame Elements) as frame-specific semantic roles with more than 100,000 annotated exemplar sentences. In addition, FrameNet defines ontological frame knowledge for each frame such as frame semantic relations, FE mappings and frame/FE definitions. The frame knowledge plays an important role in frame semantic parsing. Most previous approaches \citep{kshirsagar-etal-2015-frame, yang-mitchell-2017-joint,peng-etal-2018-learning} only use exemplar sentences and ignore the ontological frame knowledge. Recent researches \citep{jiang-riloff-2021-exploiting,su-etal-2021-knowledge} introduce frame semantic relations and frame definitions into the subtask frame identification. Different from previous work, we construct a heterogeneous graph named Frame Knowledge Graph (FKG) based on frame knowledge to model multiple semantic relations between frames and frames, frames and FEs, as well as FEs and FEs. Furthermore, we apply FKG to all subtasks of frame semantic parsing, which can fully inject frame knowledge into frame semantic parsing. The knowledge-enhanced representations of frames and FEs are learned in a unified vector space and this can also strengthen interactions between frame identification and other subtasks.

Most previous systems neglect interactions between subtasks, they either focus on one or two subtasks \citep{hermann-etal-2014-semantic,fitzgerald-etal-2015-semantic,marcheggiani-titov-2020-graph} of frame semantic parsing or treat all subtasks independently \citep{das-etal-2014-frame, peng-etal-2018-learning}. Furthermore, in argument identification and role classification, previous approaches process each argument separately with sequence labeling strategy \citep{yang-mitchell-2017-joint, bastianelli2020encoding} or span-based graphical models \citep{tackstrom-etal-2015-efficient, peng-etal-2018-learning}. In this paper, we propose Frame Semantic Graph (FSG) to represent frame semantic structures and treat frame semantic parsing as a process to construct this graph incrementally. With graph structure, historical decisions of parsing can guide the current decision of argument identification and role classification, which highlights interactions between subtasks and arguments.

Based on two graphs mentioned above, we propose our framework \textsc{Kid} (\uline{\textbf{K}}nowledge-guided \uline{\textbf{I}}ncremental semantic parser with \uline{\textbf{D}}ouble-graph). FKG provides a static knowledge background for encoding frames and FEs while FSG represents dynamic parsing results in frame semantic parsing and highlights relations between arguments. 

Overall, our contributions are listed as follow:
\begin{itemize}
    \item  We build FKG based on the ontological frame knowledge in FrameNet. FKG incorporates frame semantic parsing with structured frame knowledge, which can get knowledge-enhanced representations of frames and FEs.
    \item We propose FSG to represent the frame semantic structures. We treat frame semantic parsing as a process to construct the graph incrementally. This graph focuses on the target-argument and argument-argument relations.
\end{itemize}

We evaluate the performance of \textsc{Kid} on two FrameNet datasets: FN 1.5 and FN 1.7, the results show that the \textsc{Kid} achieves state-of-the-art on these datasets by increasing up to 1.7 points on F1-score. Our extensive experiments also verify the effectiveness of these two graphs. 

\section{Ontological Frame Knowledge}
Frame semantics relates linguistic semantics to encyclopedic knowledge and advocates that one cannot understand the semantic meaning of one word without essential frame knowledge related to the word \citep{fillmore2001frame}. The frame knowledge of a frame contains frame/FE definitions, frame semantic relations and FE mappings. FrameNet defines 8 kinds of frame semantic relations such as \textbf{Inheritance}, \textbf{Perspective\_on} and \textbf{Using}; for any two related frames, the FrameNet defines FE mappings between their FEs. For example, the frame \textbf{\textit{Receiving}} inherits from \textbf{\textit{Getting}} and the FE \textbf{Donor} of \textbf{\textit{Receiving}} is mapped to the FE \textbf{Source} of \textbf{\textit{Getting}}. Each frame or FE has its own definition and may mention other FEs.

We propose two ways of reasoning about frame semantic parsing: inter-frame reasoning and intra-frame reasoning in Figure \ref{fig2}. Frame knowledge mentioned above can guide both ways of reasoning. The frame semantic relation between \textbf{\textit{Receiving}} and \textbf{\textit{Getting}} and FE mappings associated with it allow us to learn from the left sentence when parsing the right sentence because similar argument spans of two sentences will have related FEs as their roles. The FE definitions reflect dependencies between arguments. The definition of \textbf{Role} in frame \textbf{\textit{Receiving}} mentions \textbf{Theme} and \textbf{Donor}, which reflects dependencies between argument \textit{the book} and argument \textit{as a gift}.

\section{Task Formulation}
Given a sentence $S=w_0,\dots, w_{n-1}$ with a target span $t$ in $S$, the frame semantic parsing aims to extract the frame semantic structure of $t$. Suppose that there are $k$ arguments of $t$ in $S$: $a_0,\dots,a_{k-1}$, subtasks can be formulated as follow:
\begin{itemize}
    \item Frame identification: finding an $f\in \mathcal{F}$ evoked by target $t$, where $\mathcal{F}$ denotes the set of all frames in the FrameNet.
    \item Argument identification: finding the boundaries $i^s_\tau$ and $i^e_\tau$ for each argument $a_\tau=w_{i^s_\tau},\dots,w_{i^e_\tau}$.
    \item Role classification: assigning an FE $r_\tau \in \mathcal{R}_f$ to each $a_\tau$, where $\mathcal{R}_f$ denotes the set of all FEs of frame $f$.
\end{itemize}

\section{Method}

\textsc{Kid} encodes all frames and FEs to knowledge-enhanced representations via frame knowledge graph encoder (section \ref{sec: knowledge}). For a sentence with a target, contextual representations of tokens are derived from the sentence encoder (section \ref{sec: sentence}). Frame semantic parsing is regarded as a process to build FSG incrementally from an initial target node to complete FSG. Frame identification finds a frame evoked by the target and combines the target with its frame into the initial node of FSG (section \ref{sec: fi}). Argument identification (section \ref{sec: arg}) and role classification (section \ref{sec: role}) for each argument is based on the current snapshot of partial FSG considering all historical decisions. Section \ref{sec: fsg} tells how frame semantic graph decoder encodes partial FSG to its representation and how it expands FSG incrementally, which is also shown in Figure \ref{fig4}. 

\subsection{Frame Knowledge Graph Encoder}
\label{sec: knowledge}

\begin{figure}[t]
\centering
\includegraphics[width=0.48\textwidth]{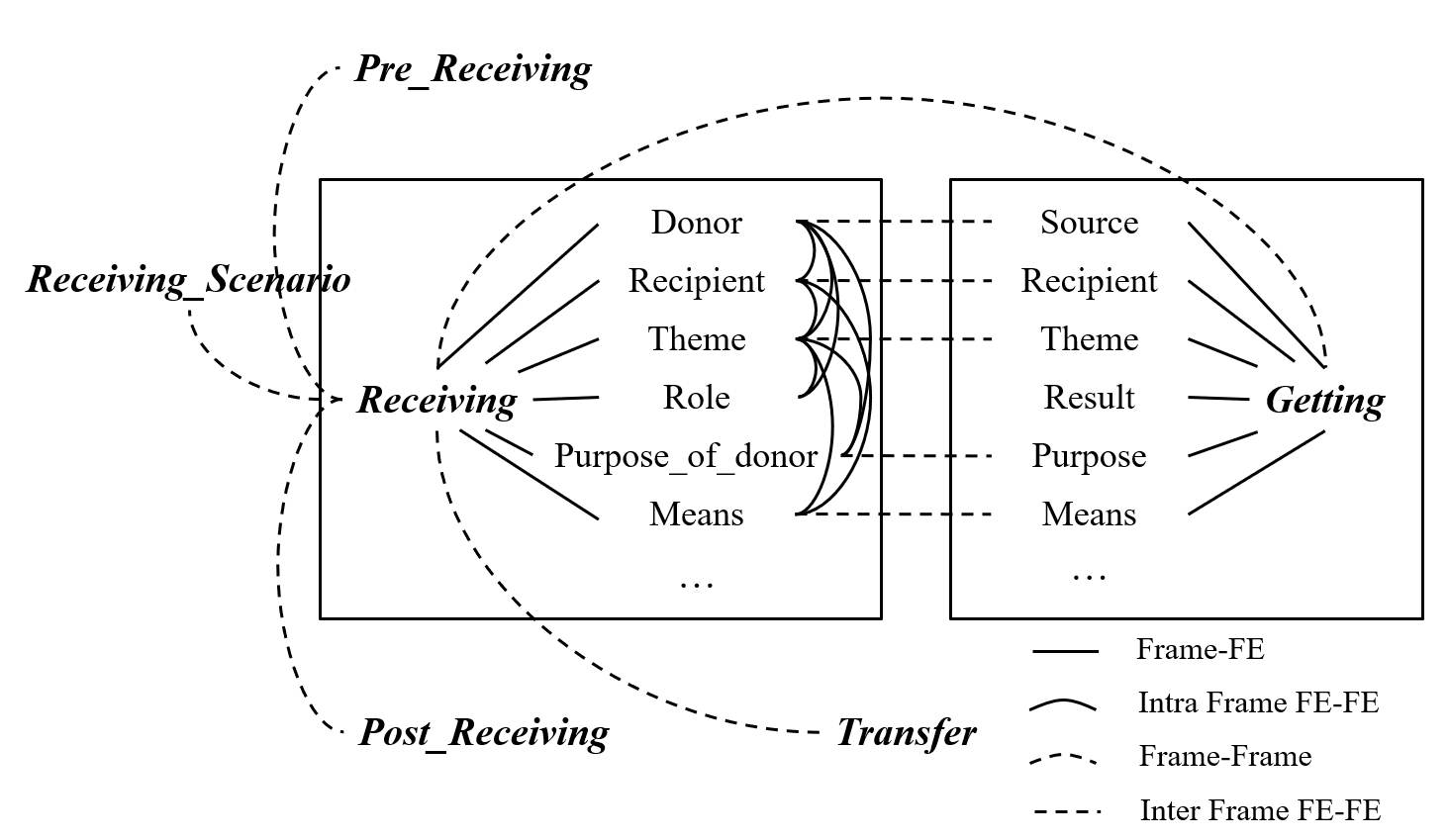} 
\caption{A subgraph of FKG. We only show intra-frame and inter-frame FE-FE relations in the frame \textbf{\textit{Receiving}}. Inside the solid rectangular box are a frame and its FEs.}
\label{fig3}
\end{figure}

FKG is an undirected multi-relational heterogeneous graph, and Figure \ref{fig3} shows a subgraph of FKG. Its nodes contain both frames and FEs and there are four kinds of relations in FKG: frame-FE, frame-frame, inter-frame FE-FE and intra-frame FE-FE relations. The following will show how we extract these relations from frame knowledge:

\noindent\textbf{Frame-FE}: we connect a frame with its FEs so that we can learn representations of frames and FEs in a unified vector space to strengthen interactions between frame identification and other subtasks.

\noindent\textbf{Frame-frame} and \textbf{inter-frame FE-FE}: these two kinds of relations are frame semantic relations and FE mappings respectively and here we ignore relation types of frame semantic relations. They can both guide inter-frame reasoning in Figure \ref{fig2}.

\noindent\textbf{Intra-frame FE-FE}: If the definition of an FE mentions another FE in the same frame, they will have intra-frame FE-FE relations with each other. This relation can help with intra-frame reasoning and strengthen interactions between arguments.

\begin{figure*}[t]
\centering
\includegraphics[width=1\textwidth]{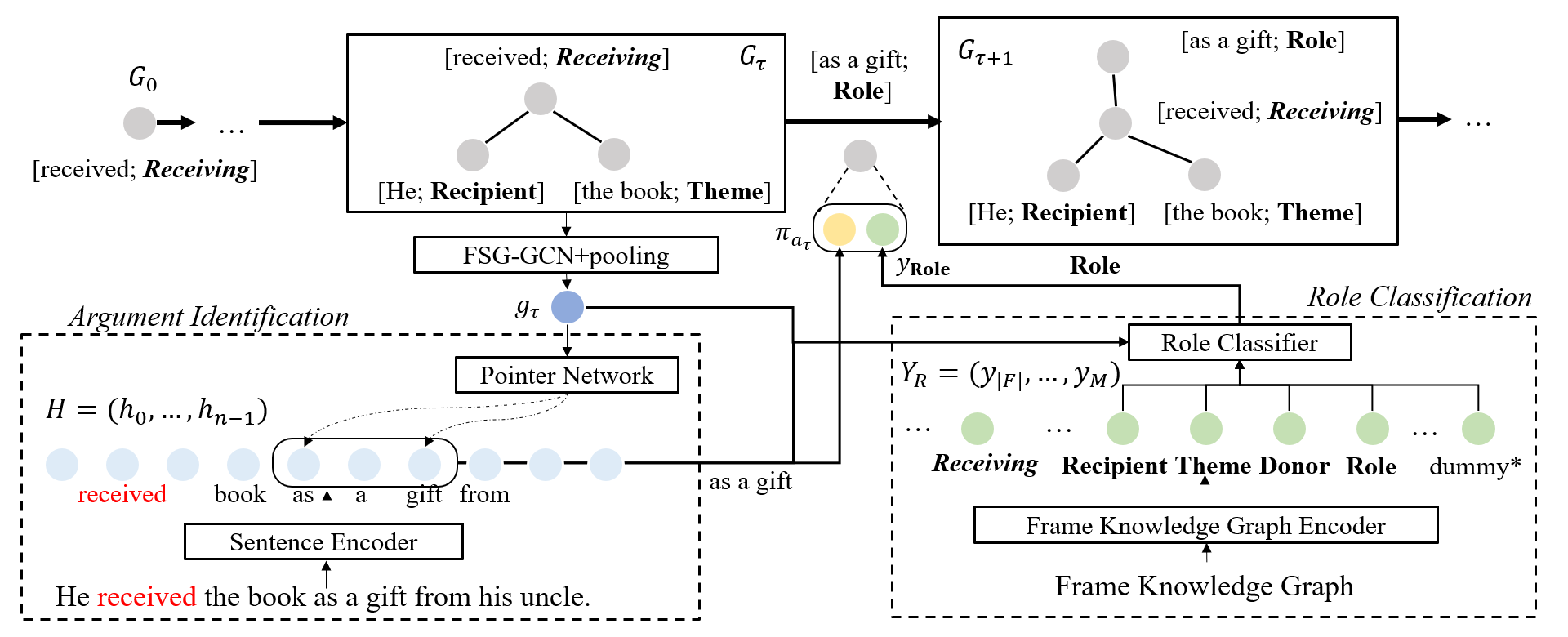} 
\caption{Based on the representation $g_\tau$ of partial $G_\tau$, frame semantic graph decoder identifies new argument \textit{as a gift} with pointer networks, and label it with FE \textbf{Role}. $G_\tau$ will be updated to $G_{\tau+1}$ with (\textit{as a gift}, \textbf{Role}).} 
\label{fig4}
\end{figure*}

The frame knowledge graph encoder aims to get knowledge-enhanced representations of nodes in FKG via an RGCN \citep{schlichtkrull2018modeling} module. We use $\mathcal{F}$ to represent all frames in FrameNet and $\mathcal{R}_f$ to represent all FEs of frame $f$. In addition, we use $\mathcal{R}= \bigcup_{f\in \mathcal{F}} \mathcal{R}_f$ to represent all FEs in the FrameNet. Let $0, \dots, |\mathcal{F}| - 1$ denote all frames and $|\mathcal{F}|, \dots, |\mathcal{F}| + |\mathcal{R}| - 1$ denote all FEs. Moreover, we introduce a special dummy node indexing $|\mathcal{F}| + |\mathcal{R}|$ into FKG. So the vectors $y_0,\dots,y_{M} \in \mathbb{R}^{d_n}$ denote the representations of all nodes in FKG, where $M = |\mathcal{F}| + |\mathcal{R}|$. 

For each node $i$, we take a randomly initialized embedding $y^{(0)}_i \in \mathbb{R} ^{d_n}$ as the input feature of the RGCN module. Then we can get representations of all frames and FEs:
\begin{equation}
    y_0,\dots,y_M = \mathrm{RGCN}\left(y^{(0)}_0, \dots, y^{(0)}_M\right)
\end{equation}

The RGCN module models four kinds of relations: Frame-FE, intra-frame FE-FE, frame-frame and inter-frame FE-FE. For better modeling inter-frame relations between FEs, we also fuse name information into representations of FEs. The FEs whose names are the same will share the same embeddings, i.e. for $i, j \ge |\mathcal{F}|$, $y^{(0)}_i=y^{(0)}_j$ if the name of $i$ is the same as $j$.

\subsection{Sentence Encoder}
\label{sec: sentence}
The sentence encoder converts tokens of the sentence $S=w_0,\dots,w_{n-1}$ to their representations $h_0,\dots, h_{n-1} \in \mathbb{R}^{d_h}$.

We use LSTM \citep{hochreiter1997long} and GCN \citep{kipf2016semi} to model both sequential structure and dependency structure:
\begin{align}
\alpha_0,\dots,\alpha_{n-1} &= \mathrm{BiLSTM}\left(e_0,\dots,e_{n-1}\right) \\
    \beta_0, \dots, \beta_{n-1} &=   \mathrm{GCN}\left(\alpha_0,\dots,\alpha_{n-1}, \mathcal{T}
    \right) 
\end{align}

$e_i$ denotes the embedding of word $w_i$. We get contextual representations $h_i$ by adding $\beta_i$ to $\alpha_i$. We follow previous studies \citep{marcheggiani-titov-2020-graph, bastianelli2020encoding} to use syntax structures like dependency tree $\mathcal{T}$ of $S$ here.

Furthermore, we use boundary information \citep{wang-chang-2016-graph, cross2016span, ouchi-etal-2018-span} to represent spans like $s = w_i,\dots,w_j$ based on token representations because we need to embed spans into the vector space of FKG in frame identification and role classification:
\begin{equation}
    Q(i,j) = \mathrm{FFN}\left( (h_j-h_i)\oplus(h_j+h_i)\right)
\end{equation}

The dimension of $Q(i,j)$ is $d_n$. The $\oplus$ denotes concatenation operation and FFN denotes Feed Forward Network.

\subsection{Frame Semantic Parsing}
\subsubsection{Frame Identification}
\label{sec: fi}
A frame $f \in \mathcal{F}$ will be identified based on the target $t$, representations of tokens $h_0,\dots,h_{n-1}$ and representations of frames $y_0,\dots,y_{|\mathcal{F}|-1}$ with a scoring module.
The target $t=w_{i_t^s},\dots,w_{i_t^e}$ will be embedded to the vector space of all frames as $\gamma_t \in \mathbb{R} ^{d_n}$. We can calculate dot product normalized similarities between $\gamma_t$ and all frames $Y_{\mathcal{F}} = (y_0,\dots,y_{|\mathcal{F}|-1}) \in \mathbb{R}^{d_n \times |\mathcal{F}|}$ to get the probability distribution of $f$. For short, we use $\pi_t$ to denote $Q(i^s_t,i^e_t)$:
\begin{align}
    \gamma_t &= \mathrm{tanh}\left( \mathrm{FFN}(\pi_t)\right)\\
    P\left(f|S,t\right) &= \mathrm{softmax}\left(Y_{\mathcal{F}}^\top\cdot \gamma_t\right)
\end{align} 

\subsubsection{Argument Identification}
\label{sec: arg}
Based on $g_\tau$, the representation of current snapshot of FSG $G_\tau$, we need to find an argument $a_\tau = w_{i^s_\tau},\dots,w_{i^e_\tau}$. We use pointer networks \citep{vinyals2015pointer} to identify its start and end positions $i^s_\tau$ and $i^e_\tau$ separately via an attention mechanism, which is more efficient than traditional span-based model \citep{chen-etal-2021-joint}. Take $i^s_\tau$ as example:
\begin{align}
    \rho^{s}_\tau &= \mathrm{FFN} \left( g_\tau\right) \\
    P\left(i^s_\tau|S,G_\tau\right)&=\mathrm{softmax}\left(H^\top\cdot \rho^{s}_\tau\right) 
\end{align}

$H = (h_0,\dots,h_{n-1}) \in \mathbb{R}^{d_h\times n}$ represents the output of the sentence encoder, and $\rho^s_\tau \in \mathbb{R}^{d_h}$ is used to find the start position of argument span $a_\tau$.

\subsubsection{Role Classification}
\label{sec: role}
Based on $g_\tau$ and $a_\tau$, we embed $a_\tau$ into the vector space of FEs as $\gamma_{a_\tau} \in \mathbb{R}^{d_n}$. Similar to frame identification, we calculate dot product normalized similarities between $\gamma_{a_\tau}$ and all FEs $Y_{\mathcal{R}} = (y_{|\mathcal{F}|},\dots,y_{|\mathcal{F}| + |\mathcal{R}|}) \in \mathbb{R}^{d_n \times (|\mathcal{R}|+1)}$ to get the conditional probability distribution of $r$ given $a_\tau$ and $G_\tau$.
\begin{align}
    \gamma_{a_\tau} &= \mathrm{FFN}(\pi_{a_\tau} \oplus g_\tau)\\
    P\left(r_\tau|S,G_\tau,a_\tau\right) &= \mathrm{softmax}\left(Y_{\mathcal{R}}^\top\cdot \gamma_{a_\tau}\right) 
\end{align}

\subsection{Frame Semantic Graph Decoder}
\label{sec: fsg}

We propose FSG to represent the frame semantic structure of $t$ in the sentence $S$ and we treat the frame semantic parsing as a process to construct FSG incrementally. Intermediate results of FSG are partial FSGs representing all historical decisions, which highlights interactions between arguments. Suppose that there are $k$ arguments of target $t$: $a_0,\dots,a_{k-1}$ with their roles $r_0, \dots, r_{k-1}$. For $\tau$-th snapshot of FSG $G_\tau$, it contains $\tau+1$ nodes: one target node $(t,f)$ and $\tau$ argument nodes (if exist) $(a_0,r_0),\dots,(a_{\tau-1},r_{\tau-1})$. The target node will be connected with all argument nodes. The indices of nodes in $G_\tau$ depend on the order in which they are added into the graph, 0 denotes the target node and $1,\dots,\tau$ denotes $(a_0,r_0),\dots,(a_{\tau-1},r_{\tau-1})$. 

We encode $G_\tau$ to its representation $g_\tau$:
\begin{gather}
    g_\tau = \mathrm{Maxpooling}\left(z_0,\dots,z_\tau\right)\\
    z_0,\dots,z_\tau = \mathrm{GCN}\left(z^{(0)}_0,\dots,z^{(0)}_\tau, G_\tau\right)\\
    z^{(0)}_j=\left\{
             \begin{array}{ll}
             \pi_t\oplus y_{i_f}, &  j=0 \\
             \pi_{a_j} \oplus y_{i_{r_j}}, & j=1,\dots,\tau \\
             \end{array}
    \right. 
\end{gather}

\noindent where $i_f$ and $i_{r_j}$ denotes indices of $f$ and $r_j$ in FKG, and $\pi_{a_j} = Q(i^s_j,i^e_j)$. The GCN module is to encode partial FSG.

Based on the representation $g_\tau$ of each snapshot $G_\tau$, \textsc{Kid} predicts boundary positions of argument $a_\tau$ and assign an FE $r_\tau$ as its semantic role (section \ref{sec: arg},\ref{sec: role}). The $G_\tau$ will be updated to $G_{\tau+1}$ with the new node $(a_\tau, r_\tau)$ until the $r_\tau$ is the special dummy node in FKG. Figure \ref{fig4} shows how to find a new node and add it into the FSG.

\section{Training and Inference}

\subsection{Training}

We train \textsc{Kid} with all subtasks jointly by optimizing the loss function $\mathcal{L}$ since representations of frames and FEs are learned in a unified vector space.

\begin{gather}
\mathcal{L}^{f} = -\log P(f = f^{\mathrm{gold}}|S,t)\\
\mathcal{L}^{a}_{s/e} = -\sum_{\tau=0}^{k-1}\log P(i^{s/e}_\tau = I^{s/e}_{\tau}|S,G_\tau)
\\
\mathcal{L}^{r} = -\sum_{\tau=0}^{k}\log P(r_\tau=r^{\mathrm{gold}}_\tau|S,G_\tau,a_\tau)
\\
\mathcal{L} = \lambda_1\mathcal{L}^{f} + \lambda_2(\mathcal{L}^{a}_s +\mathcal{L}^{a}_e) +\lambda_3\mathcal{L}^{r}
\end{gather}

\noindent where $f^{\mathrm{gold}}$ is the gold frame and $ I^s_{\tau},I^e_\tau,r^{\mathrm{gold}}_\tau$ are gold labels of argument $a_\tau$. $r_k^{\mathrm{gold}}$ is ``Dummy'', indicating the end of the parsing. We force our model to identify arguments in a left-to-right order, i.e. $a_0$ is the leftmost argument in $S$. We use gold frame in the initial node of FSG: $G_0 = (t,f^{gold})$ while other nodes are predicted autoregressively: $G_{\tau+1} = G_\tau + (\hat{a}_\tau,\hat{r}_\tau)$, $\hat{r}_\tau \in \mathcal{R}_{f^{\mathrm{gold}}}$.

\subsection{Inference}

\textsc{Kid} predicts frame and all arguments with their roles in a sequential way. We use probabilities above with some constraints: 1. We use lexicon filtering strategy: for a target $t$, we can use the lemma $\ell_t$ of it to find a subset of frames $\mathcal{F}_{
\ell_t} \subset \mathcal{F}$ so that we can reduce the searching space; 2. Similarly, we take $\mathcal{R}_{\hat{f}}$ instead of $\mathcal{R}$ as the set of candidate FEs; 3. In argument identification, we will mask spans that are already selected as arguments, and $i^e_\tau$ should be no less than $i^s_\tau$.

\section{Experiment}

\begin{table}
\centering\small
\begin{tabular}{lcccc}
\toprule
             & \#exemplar & \#train & \#dev & \#test \\ \midrule
FN 1.5 & 153952     & 17143   & 2333  & 4458\\
FN 1.7 & 192461     & 19875   & 2309  & 6722\\ \bottomrule
\end{tabular}
\caption{Number of instances in two datasets.
}
\label{tab1}
\end{table}

\begin{table*}[t]
\centering\small
\begin{tabular}{lccccccc}
 \toprule
 \multirow {2} {*} {Model}& \textbf{Frame Id} &
  \multicolumn{3}{c}{\textbf{Arg Id (gold frame)}} &
  \multicolumn{3}{c}{\textbf{Full structure}}\\
                 & Accuracy & Precision             & Recall             & F1-score   & Precision & Recall &F1-score         \\ \midrule
SEMAFOR \citeyearpar{das-etal-2014-frame}     &    83.6 & 65.6          & 53.8          & 59.1     &- &- &-             \\\citet{hermann-etal-2014-semantic}   &88.4& -&-&-             & 74.3          & 66.0          & 69.9          \\
SEMAFOR (HI)
\citeyearpar{kshirsagar-etal-2015-frame}& - & 67.2          & 54.8          & 60.4 &- &- &-                       \\

\citet{tackstrom-etal-2015-efficient} &-  &-&-&-            & \textbf{75.4}          & 65.8          & 70.3          \\
\citet{fitzgerald-etal-2015-semantic}  &-&-&-&-          & 74.8          & 65.5          & 69.9          \\
open-SESAME \citeyearpar{Swayamdipta2017FrameSemanticPW}&86.9   & \textbf{69.4}          & 60.5          & 64.6            & 71.0          & 67.8          & 69.4                    \\
\textsc{Kid} (GloVe) & \textbf{89.5} & 64.6 & \textbf{68.2} & \textbf{66.4} & 73.8 & \textbf{76.8} & \textbf{75.3} 
\\ \midrule SEMAFOR (HI + exemplar) \citeyearpar{kshirsagar-etal-2015-frame}& -&66.0          & 60.4          & 63.1           &-&-&-           \\\citet{yang-mitchell-2017-joint} &88.2& \textbf{70.2} & 60.2 & 65.5 & 77.3          & 71.2          & 74.1           \\ \citet{swayamdipta-etal-2018-syntactic}& -&67.8          & 66.2          & 67.0&-&-&- 
\\ 

\citet{peng-etal-2018-learning}&89.2 &-&-&-                   & \textbf{79.2} & 71.7          & 75.3         \\\citet{marcheggiani-titov-2020-graph}&-& 69.8 & 68.8          & 69.3                    &-&-&- 
\\
\citet{chen-etal-2021-joint}        &89.4 &-&-&-          & 75.1          & 76.9          & 76.0          \\
\textsc{Kid} (GloVe + exemplar) & \textbf{90.0}& 66.8          & \textbf{73.7}          & \textbf{70.1}            & 75.5          & \textbf{80.1} & \textbf{77.7}   \\ \midrule
\citet{bastianelli2020encoding} (JL)   &90.1 & \textbf{74.6}          & 74.4          & 74.5      &-&-&-  
\\
\citet{chen-etal-2021-joint} (BERT)    & 90.5  &-&-&-         & 78.2          & 82.4          & 80.2         \\ \citet{jiang-riloff-2021-exploiting}&91.3&-&-&-&-&-&- \\
\citet{su-etal-2021-knowledge}&\textbf{92.1}&-&-&-&-&-&-\\
\textsc{Kid} (BERT)&91.7&71.7 &\textbf{79.0} &\textbf{75.2}        & \textbf{79.3} & \textbf{84.2} & \textbf{81.7}  \\ \bottomrule            
\end{tabular}
\caption{Empirical results on the test set of FN 1.5. All models are single-task, non-ensemble. The upper block lists models trained without exemplar instances, the lower block lists models with pretrained language models. \textsc{Kid} outperforms other models under all conditions except \citet{su-etal-2021-knowledge}. We also train our model with multiple runs and report the statistical analysis with a significance testing in appendix \ref{sec: exp}.}
\label{tab2}
\end{table*}

\subsection{Datasets}
We evaluate \textsc{Kid} on two FrameNet datasets: FN 1.5 and FN 1.7.\footnote{\url{https://framenet.icsi.berkeley.edu/fndrupal/about}} FN 1.7 is an extension version of FN 1.5, including more fine-grained frames and more instances. FN 1.5 defines 1019 frames and 9634 FEs while FN 1.7 defines 1221 frames and 11428 FEs. We use the same splits of datasets as \citet{peng-etal-2018-learning}, and we also follow \citet{kshirsagar-etal-2015-frame, yang-mitchell-2017-joint,peng-etal-2018-learning,chen-etal-2021-joint} to include exemplar instances as training instances. As \citet{kshirsagar-etal-2015-frame} states that there exists a domain gap between exemplar instances and original training instances, we follow \citet{chen-etal-2021-joint} to use exemplar instances as pre-training instances and further train our model in original training instances. Table \ref{tab1} shows the numbers of instances in two datasets.

\subsection{Empirical Results}

We compare \textsc{Kid} with previous models (see appendix \ref{sec: exp}) on FN 1.5 and FN 1.7. We focus on three metrics: frame acc, arg F1 and full structure F1.\footnote{\url{https://www.cs.cmu.edu/~ark/SEMAFOR/eval/}} Full structure F1 shows the performance of models on extracting full frame semantic structures from text, frame acc denotes accuracy of frame identification and arg F1 evaluates the results of argument identification and role classification with gold frames. All metrics are evaluated in test set.

Table \ref{tab2} shows results on FN 1.5. For a fair comparison, we divide models into three parts: the first part of models do not use exemplar instances as training data; the second part of models use exemplar instances without any pretrained language models; the third part of models use pretrained language models. \textsc{Kid} (GloVe) uses GloVe \citep{pennington-etal-2014-glove} as word embeddings and \textsc{Kid} (BERT) fine-tunes pretrained language model BERT \citep{devlin-etal-2019-bert} to encode word representations. \textsc{Kid} achieves state-of-the-art of these metrics under almost all circumstances, and we also train our model with multiple runs, which shows \textsc{Kid} (GloVe + exemplar) and \textsc{Kid} (BERT) outperforms  previous state-of-the-art models by 1.4 and 1.3 full structure F1-score averagely. There is an exception that our model with BERT does not outperform \citet{su-etal-2021-knowledge} on frame identification accuracy and we find that the number of train, validation and test instances reported by them are a little bit smaller than ours. Results on FN 1.7 and statistical analysis of our model with multiple runs are listed in appendix \ref{sec: exp}.

It is worth noting that \textsc{Kid} achieves much higher recall than other models. We attribute this to the incremental strategy of building FSG. By constructing FSG incrementally, \textsc{Kid} can capture relations between arguments and identify arguments that are hard to find in other models.

\subsection{Ablation Study}

To prove the effectiveness of double-graph architecture, we conduct further experiments with \textsc{Kid} on FN 1.5. Table \ref{tab3} shows ablation study on double-graph architecture. w/o FSG uses LSTM instead of our frame semantic graph decoder. It takes a sequence of arguments and their roles that have already been identified as input to predict the next argument. FSG performs better than LSTM because it captures target-argument and argument-argument relations and can model long-distance dependencies. w/o FKG directly uses input vectors of frame knowledge graph encoder, and results also show that knowledge-enhanced representations are better than randomly initialized embeddings. We also test the influence of double-graph structure with pre-trained language models, the results shows the double-graph structure is still effective and useful even with the pre-trained language models.

FKG is a multi-relational heterogeneous graph. The ablation study on structures of FKG is shown in Table \ref{tab4}. In addition, we evaluate the performance of FI w/o FKG, which identifies frames with a simple linear classification layer instead of FKG, and the results prove that FKG strengthens interactions between frame identification and role classification.

\begin{table}[]
    \centering\small
    \begin{tabular}{lcc}
    \toprule
    Model & Full structure F1 & Arg F1  \\
    \midrule
         \textsc{Kid} (GloVe) & 75.28 & 66.35 \\
         w/o FSG& 74.43 & 64.99 \\
         w/o FKG & 74.60 & 64.96\\
         w/o double-graph &73.34 & 63.41\\
         \midrule
         \textsc{Kid} (BERT w/o exemplar)  & 79.44 & 71.59 \\
         w/o FSG & 78.84 & 70.63\\
         w/o FKG & 79.29 & 70.86\\
         w/o double-graph & 77.77 & 68.77 \\
    \bottomrule
    \end{tabular}
    \caption{Ablation study on double-graph architecture. w/o denotes ``without''. w/o FSG uses LSTM as its decoder and w/o FKG does not use RGCN to encode frames and FEs. We also test the influence of double-graph architecture for \textsc{Kid} (BERT).}
    \label{tab3}
\end{table}

\begin{table}[]
    \centering\small
    \begin{tabular}{lcc}
    \toprule
    Model & Full structure F1 & Arg F1  \\
    \midrule
         \textsc{Kid} (GloVe) & 75.28 & 66.35 \\
         w/o frame-FE& 74.84 & 65.70 \\
         w/o frame-frame & 74.97 & 66.06\\
         w/o intra-frame FE-FE &75.10 & 66.60\\
         w/o inter-frame FE-FE &75.13 & 65.87\\
         \midrule
         FI w/o FKG & 75.00 & 65.61 \\
         w/o FKG & 74.60 & 64.96 \\
    \bottomrule
    \end{tabular}
    \caption{Ablation study on structures of FKG. We remove each kind of relations of FKG and all get a drop of full structure F1. FI w/o FKG denotes not using FKG in frame identification (FI). w/o FKG uses input vectors of frame knowledge encoder directly.}
    \label{tab4}
\end{table}

In addition, we explore the effectiveness of name information of FEs. Whether the name information is used in previous work is unclear and some BIO-based approaches like \citet{marcheggiani-titov-2020-graph} are likely to use name information by regarding FEs with the same name as the same label in role classification. To the best of our knowledge, we are the first one to study the effectiveness of name information. As shown in Table \ref{tab5}, the names of FEs provide rich information for frame semantic parsing and shared embedding strategy can make good use of the name information. Further ablation study of FKG is conducted under the circumstance without name information, the performance will drop 0.9 points if we remove FKG too, showing that knowledge-enhanced representations are important no matter whether we share embeddings for FEs with the same names or not.

\begin{table}[]
    \centering\small
    \begin{tabular}{lcc}
    \toprule
    Model & Full structure F1 & Arg F1  \\
    \midrule
         \textsc{Kid} (GloVe) & 75.28 & 66.35 \\
         w/o FKG & 74.60 & 64.96 \\
         w/o name information& 74.46 & 64.86 \\
         w/o both & 73.53 & 64.01 \\
    \bottomrule
    \end{tabular}
    \caption{Ablation study on name information. Sharing embeddings for FEs whose names are the same are quite useful. The performance will drop a lot if we remove both name information and FKG, which reveals the importance of FKG.}
    \label{tab5}
\end{table}

\begin{table}[]
    \centering\small
    \begin{tabular}{lccccc}
    \toprule
    \multirow{2}{*}{Model} & \multicolumn{5}{c}{$K$}  \\ \cmidrule{2-6}
    & 0 & 4 & 16 & 32 & full\\
    \midrule
         \textsc{Kid} (GloVe) & 56.26 & 63.28 & 65.32 & 65.95 & 70.32 \\
         w/o FKG & 0.00 & 50.70 &  56.40 & 57.59 & 63.94\\\midrule
         $\Delta$ & 56.26 & 12.58 & 8.92&8.36 &6.38\\
    \bottomrule
    \end{tabular}
    \caption{Experiments on confirming transfer learning ability of FKG. $K$ denotes the number of instances of each frame in training set. Full means adding all instances of these frames except those including target \textit{get} in train and development sets. Lack of labeled instances has much less impact on Arg F1 performance of \textsc{Kid} with FKG, which confirms our assumption.}
    \label{tab6}
\end{table}
\subsection{Transfer learning ability of FKG}
As we have discussed in Figure \ref{fig2}, if frame B is related to frame A, a sentence with frame A can contribute to parsing another sentence with frame B by inter-frame reasoning. Frame-frame and inter-frame FE-FE relations of FKG can guide \textsc{Kid} to learn experience from other frames. 

To confirm that FKG has ability of transfer learning, we design zero (few)-shot learning experiments on FN 1.7. Target word \textit{get} can evoke multiple frames in FrameNet, and we choose instances including target \textit{get} with three frames (\textbf{\textit{Arriving}}, \textbf{\textit{Getting}} and \textbf{\textit{Transition\_to\_state}}) as test instances. We remove all instances with target \textit{get} from train and development instances, and can selectively add few (or zero) instances including other targets with these three frames into train and development sets. We then compare the performance of \textsc{Kid} with \textsc{Kid} w/o FKG under zero-shot and few-shot circumstances. If FKG has ability of transfer learning, \textsc{Kid} with FKG can learn experience from other related frames like \textbf{\textit{Receiving}} and its performance will not be influenced so much by the sparsity of labels.

Table \ref{tab6} shows the results of our experiments. $K=0$ indicates zero-shot learning while $K=\{4,16,32\}$ indicates few-shot learning. \textsc{Kid} without FKG performs much worse in zero-shot learning. As the number of instances that can be seen in training grows up, the performance of \textsc{Kid} with FKG gets a steady increase while the performance of \textsc{Kid} without FKG increases rapidly. Results verify our assumption that even with few train instances FKG can guide inter-frame reasoning with its structure and allow models to learn experience from other seen frames.

\section{Related Work}
Frame semantic parsing has caught wide attention since it was released on SemEval 2007 \citep{baker-etal-2007-semeval}. The task is to extract frame structures defined in FrameNet \citep{baker-etal-1998-berkeley-framenet} from text. From then on, a large amount of systems are applied on this task, ranging from traditional machine learning classifiers \citep{johansson-nugues-2007-lth, das-etal-2010-probabilistic} to fancy neural models like recurrent neural networks \citep{yang-mitchell-2017-joint,Swayamdipta2017FrameSemanticPW} and graph neural networks \citep{marcheggiani-titov-2020-graph, bastianelli2020encoding}.

A lot of previous systems neglect interactions between subtasks and relations between arguments. They either focus on one or two subtasks \citep{hermann-etal-2014-semantic,fitzgerald-etal-2015-semantic,marcheggiani-titov-2020-graph} of frame semantic parsing or treat all subtasks independently \citep{das-etal-2014-frame, peng-etal-2018-learning}. \citet{tackstrom-etal-2015-efficient} propose an efficient global graphical model, so they can enumerate all possible argument spans and treat the assignment as the Integer Linear Programming problem. Later systems like \citet{fitzgerald-etal-2015-semantic,peng-etal-2018-learning} follow this method. \citet{Swayamdipta2017FrameSemanticPW,bastianelli2020encoding} use sequence-labeling strategy, and \citet{yang-mitchell-2017-joint} integrate these two methods with a joint model. Only few approaches like \citet{chen-etal-2021-joint} model interactions between subtasks, which use the encoder-decoder architecture to predict arguments and roles sequentially. However, the sequence modeling of \citet{chen-etal-2021-joint} does not consider structure information and is not good at capturing long-distance dependencies. We use graph modeling to enhance structure information and strengthen interactions between target and argument, argument and argument.

Only a few systems utilize linguistic knowledge in FrameNet. \citet{kshirsagar-etal-2015-frame} use FE mappings to share information in FEs. In frame identification, \citet{jiang-riloff-2021-exploiting} encode definitions of frames and \citet{su-etal-2021-knowledge} use  frame identification and frame semantic relations. However, they do not utilize ontological frame knowledge in all subtasks while we construct a heterogeneous graph containing both frames and FEs. Besides, our model does not need extra encoders to encode definitions, which reduces parameters of the model. 

Some systems also treat constituency parsing or other semantic parsing tasks like AMR as a graph construction problem. \citet{yang2020strongly} use GCN to encode intermediate constituency tree to generate a new action on the tree. \citet{cai-lam-2020-amr} construct AMR graphs with the Transformer \citep{vaswani2017attention} architecture.

\section{Conclusion}
In this paper, we incorporate knowledge into frame semantic parsing by constructing Frame Knowledge Graph. FKG provides knowledge-enhanced representations of frames and FEs and can guide intra-frame and inter-frame reasoning. We also propose frame semantic graph to represent frame semantic structures. We regard frame semantic parsing as an incremental graph construction problem. The process to construct FSG is structure-aware and can utilize relations between arguments. Our framework Knowledge-guided Incremental semantic parser with Double-graph (\textsc{Kid}) achieves state-of-the-art on FrameNet benchmarks. However, how to utilize linguistic knowledge better is still to be resolved. Future work can focus on better modeling of ontological frame knowledge, which will be useful for frame semantic parsing and transfer learning in frame semantic parsing.

\section*{Acknowledgements}
This paper is supported by the National Science Foundation of China under Grant No.61936012, 61876004 and the National Key Research and Development Program of China under Grant No. 2020AAA0106700.

\bibliography{custom}
\bibliographystyle{acl_natbib}

\appendix
\section{Model Details}
\label{sec: model}
\subsection{Graph Convolutional Network}
Graph convolution is introduced in \citet{kipf2016semi}. A GCN layer is defined as follow:
\begin{equation}
    f(H^{(l)}, G) = \sigma\left(\tilde{D}^{-\frac 1 2}\tilde{A}\tilde{D}^{-\frac 1 2}H^{(l)}W^{(l)}\right)
\end{equation}
\noindent where $H^{(l)}$ denotes hidden representations of nodes in $l$-th layer, $\sigma$ is the non-linear activation function (e.g. $\mathrm{ReLU}$) and $W$ is the weight matrix. $\tilde{D}$ and $\tilde{A}$ are separately the degree and adjacency matrices for the graph $G$. From a node-level perspective, a GCN layer can be also formalized as follow:
\begin{equation}
    h^{(l+1)}_i = \sigma\left(\sum_{j\in N(i)}\frac 1 {c_{ji}} h^{(l)}_jW^{(l)}\right)
\end{equation}
\noindent where $N(i)$ is the set of neighbors of node $i$, and $c_{ji} = \sqrt{|N(j)|}\sqrt{|N(i)|}$.

By stacking $L$ different GCN layers, we get final GCN module $\mathrm{GCN}(H^{(0)}, G)$.

\subsection{Relational Graph Convolutional Network}
Type information of graph edges is ignored in GCN, and RGCN \citep{schlichtkrull2018modeling} is proposed to model relational data from which we can benefit to model the multi-relational graph FKG. Different edge types use different weights and only edges of the same relation type $r$ are associated with the same projection weight $W_r$. From a node's view:
\begin{equation}
    h^{(l+1)}_{i} = \sigma\left(h^{(l)}_iW^{(l)}_0 + \sum_{r\in R}\sum_{j\in N^{r}_i} \frac 1 {c_{i, r}} h^{(l)}_jW^{(l)}_r\right)
\end{equation}
\noindent where $N^r_i $ denotes the set of neighbor indices of node $i$ under relation $r \in R$ and $c_{i,r}$ is a normalization constant i.e.$|N^r_i|$. 

In \textsc{Kid}, we use $\tanh$ as activation function of RGCN for normalization because we need to calculate normalized dot product similarities between frames/FEs and target/arguments.

\subsection{Encoding Dependency Tree}
We follow previous studies \citep{marcheggiani-titov-2020-graph, bastianelli2020encoding} to use syntax structures like dependency tree $\mathcal{T}$ of $S$ in \textsc{Kid} because syntax structure is proved beneficial to semantic parsing. We use Stanza \citep{qi-etal-2020-stanza}, an open-source python NLP toolkit to parse dependency syntactic structure for instances, and depGCN \citep{marcheggiani-titov-2017-encoding} to encode the syntactic structure. We simplify depGCN by ignoring directions and labels of edges in dependency tree, which means if token $i$ is head or dependent of token $j$, we will have $A^{\mathcal{T}}_{ij}=A^{\mathcal{T}}_{ji}=1$ in adjacency matrix $A^{\mathcal{T}}$ of $\mathcal{T}$.

In addition, if we use BERT as encoder, the tokens are sub-word level and the adjacency matrix will be a little bit different. Specifically, if token $i$ is the sub-word of some word $u$, token $j$ is the sub-word of some word $v$ and $u$ is head or dependent word of $v$, we will have $A^{\mathcal{T}}_{ij}=A^{\mathcal{T}}_{ji}=1$ in adjacency matrix $A^{\mathcal{T}}$.

\section{Hyper-parameter Setting}
\begin{table}
\centering
\begin{tabular}{lc}
\toprule
Hyper-parameter             & Value \\ \midrule
batch size & 32\\
learning rate & \textbf{6e-5}/1e-4\\
lr decay & 0.6 per 30 epochs\\
optimizer & Adam\\
pretrain epochs & 10/20/\textbf{30}/40/50\\
epoch & 100\\
activation function & $\mathrm{ReLU}$\\
FFN Layers & 2\\
LSTM Layers & 2\\
GCN Layers & 1/\textbf{2}\\
$d_h$, $d_n$ & 512, 256\\
$\lambda_1$, $\lambda_2$, $\lambda_3$ & 0.1, 0.3, 0.3\\

\bottomrule
\end{tabular}
\caption{Hyper-parameter settings of \textsc{Kid} (GloVe).
}
\label{tab7}
\end{table}

\begin{table}
\centering
\begin{tabular}{lc}
\toprule
Hyper-parameter             & Value \\ \midrule
bert version & bert-base-uncased\\
batch size & 16\\
learning rate & \textbf{6e-6}/1e-5\\
optimizer & BertAdam\\
pretrain epochs & 10/20/\textbf{30}/40/50\\
epoch & 100\\
activation function & $\mathrm{ReLU}$\\
FFN Layers & 2\\
LSTM Layers & 2\\
GCN Layers & 2\\
$d_h$, $d_n$ & 512, 256\\
$\lambda_1$, $\lambda_2$, $\lambda_3$ & 0.1, 0.3, 0.3\\

\bottomrule
\end{tabular}
\caption{Hyper-parameter settings of \textsc{Kid} (BERT).
}
\label{tab8}
\end{table}
For replicability of our work, we list hyper-parameter settings of \textsc{Kid} (GloVe) and \textsc{Kid} (BERT) in Table \ref{tab7} and \ref{tab8}. We use the development set to manually tune the optimal hyper-parameters based on Full structure F1. The values of hyper-parameters finally selected are in bold. Token embeddings we use in \textsc{Kid} (GloVe) are the same as \citet{chen-etal-2021-joint}, including word, lemma and POS tag embeddings with a binary type embedding to distinguish whether a token is a target or not.

\section{Experiment Details}
\label{sec: exp}

\subsection{Models}
We compare \textsc{Kid} with following baselines:

\noindent\textbf{SEMAFOR}: a widely-used open-resource statistical model proposed by \citet{das-etal-2010-probabilistic, das-etal-2014-frame}.

\noindent\textbf{SEMAFOR (HI)}: an improved version of \textbf{SEMAFOR} using exemplar instances and hierarchy features (FE mappings) proposed by \citet{kshirsagar-etal-2015-frame}

\noindent\textbf{Hermann et al. \citeyearpar{hermann-etal-2014-semantic}}: a neural network-based model learning representations of words and frames.

\noindent\textbf{T{\"a}ckstr{\"o}m et al. \citeyearpar{tackstrom-etal-2015-efficient}}: identifying arguments with a global graphical model.

\noindent\textbf{FitzGerald et al. \citeyearpar{fitzgerald-etal-2015-semantic}}: an extension of \citet{tackstrom-etal-2015-efficient} learning neural representations of frames and FEs.

\noindent\textbf{open-SESAME}: a syntax-free open-resource semantic parser proposed by \citet{Swayamdipta2017FrameSemanticPW}.

\noindent\textbf{Swayamdipta et al. \citeyearpar{swayamdipta-etal-2018-syntactic}}: an extension version of \textbf{open-SESAME} with multi-task and exemplar instances.

\noindent\textbf{Yang and Mitchell \citeyearpar{yang-mitchell-2017-joint}}: a joint model integrating both sequential and relational models.

\noindent\textbf{Peng et al. \citeyearpar{peng-etal-2018-learning}}: a joint model using latent structure variables.

\noindent\textbf{Chen et al. \citeyearpar{chen-etal-2021-joint}}: a joint encoder-decoder model predicting arguments and roles sequentially.

\noindent\textbf{Marcheggiani and Titov \citeyearpar{marcheggiani-titov-2020-graph}}: a GCN-based model over constituency trees.

\noindent\textbf{Bastianelli et al. \citeyearpar{bastianelli2020encoding}  (JL)}: a GCN-based model encoding syntactic constituency path. JL denotes joint learning on all subtasks of frame semantic parsing.

\noindent\textbf{Kalyanpur et al. \citeyearpar{kalyanpur2020open}}: a T5-based model treating frame semantic parsing as a sequence-to-sequence generation task.

\noindent \textbf{Jiang and Riloff \citeyearpar{jiang-riloff-2021-exploiting}}: a sentence-pair bert-based model using frame definitions.

\noindent \textbf{Su et al. \citeyearpar{su-etal-2021-knowledge}}: a BERT-based model for frame identification using both frame identification and frame semantic relations.

\begin{table}
\centering\small
\begin{tabular}{lccc}
\toprule
        Model   & Precision    & Recall    & F1-score   \\ \midrule
\citet{peng-etal-2018-learning}       & 78.0 & 72.1 & 75.0 \\
\textsc{Kid} (GloVe)        & 77.0 & \textbf{79.8} & \textbf{78.4} \\ \midrule
\textsc{Kid} (BERT) & 81.1 & 83.3 & 82.2 \\ \bottomrule
\end{tabular}
\caption{Full structure F1 on FN 1.7.}
\label{tab9}
\end{table}

\begin{table}
\centering\small
\begin{tabular}{lccc}
\toprule
        Model   & Frame Acc.   \\ \midrule
\citet{peng-etal-2018-learning}       & 88.6 \\
\textsc{Kid} (GloVe)        & \textbf{89.5} \\ \midrule
\citet{jiang-riloff-2021-exploiting} & 92.1 \\
\citet{su-etal-2021-knowledge} & \textbf{92.4} \\
\textsc{Kid} (BERT) & 91.7 \\ \bottomrule
\end{tabular}
\caption{Frame Acc. on FN 1.7.}
\label{tab10}
\end{table}

\begin{table}
\centering\small
\begin{tabular}{lccc}
\toprule
        Model       & Precision    & Recall    & F1-score   \\ \midrule
open-SESAME \citeyearpar{Swayamdipta2017FrameSemanticPW}       & 62 & 55 & 58 \\
\textsc{Kid} (GloVe)       & \textsc{69.2} & \textsc{73.3} & \textsc{71.2} \\ \midrule
\citet{kalyanpur2020open} & 71 & 73 & 72 \\
\textsc{Kid} (BERT) & \textbf{74.1} & \textbf{77.3} & \textbf{75.6} \\ \bottomrule
\end{tabular}
\caption{Arg F1 on FN 1.7. Results of other models are obtained from \citet{kalyanpur2020open}.}
\label{tab11}
\end{table}

\begin{table}
\centering
\begin{tabular}{lccc}
\toprule
        Model       & Time cost (\%) \\ \midrule
\textsc{Kid} (GloVe)       & 27.17 \\ 
\textsc{Kid} (BERT) & 18.85 \\ \bottomrule
\end{tabular}
\caption{Time cost of FKG. We list the proportion of time encoding FKG in whole runtime. Encoding FKG may slightly hurt the efficiency of models but is not the bottleneck of our models.}
\label{tab12}
\end{table}

\begin{table*}[]
    \centering\small
    \begin{tabular}{llccccccc}
    \toprule
    \multirow{2}{*}{Model} & \multirow{2}{*}{Metrics} &  \multicolumn{5}{c}{seed} & \multirow{2}{*}{Avg. $\pm$ Dev.}& \multirow{2}{*}{ $p$-value}     \\ \cmidrule{3-7}
    & & $s_1$ & $s_2$ & $s_3$ & \textbf{$s_4$} & $s_5$ & \\
    \midrule
         \multirow{3}{*}{\textsc{Kid} (GloVe)} & Frame Acc. & 89.14 & 88.90 & 89.12 & 89.48 & 88.55 & 89.04 $\pm$ 0.34&0.007\\
          & Arg F1 & 65.65 &  66.30 & 65.87 & 66.35&65.74&65.98 $\pm$ 0.32& 
        *\\ & Full structure F1 & 74.76 &75.34&74.94&75.28&74.57&74.98 $ \pm$ 0.33 & * \\\midrule  \multirow{3}{*}{\textsc{Kid} (GloVe + exemplar)} & Frame Acc. & 89.74 & 89.56 & 89.75 & 89.90 & 89.93 & 89.78 $\pm$ 0.15&0.002\\
          & Arg F1 & 69.63 &  69.85 & 69.48 & 70.08&69.25&69.66 $\pm$ 0.32&0.033\\ & Full structure F1 & 77.27 &77.41&77.36&77.73&77.32&77.42 $\pm$ 0.18&* \\ \midrule
         \multirow{3}{*}{\textsc{Kid} (BERT)} & Frame Acc. & 91.81 &	91.92 &	91.81 &	91.74	 &91.63
 & 91.78 $\pm$ 0.11&-\\
          & Arg F1 & 74.76 &	74.55 &	74.98 &	75.17 &	74.60 &
74.81 $\pm$ 0.26&0.028\\ & Full structure F1 & 81.50&81.44&81.61&81.68&81.38&81.52 $\pm$ 0.13&*\\
    \bottomrule
    \end{tabular}
    \caption{Statistical analysis of multiple runs on FN 1.5. We train our model with five different random seeds $s_1-s_5$ and the results with seed $s_4$ are reported in Table \ref{tab2}. We both report the average performance with deviation and the results of significance testing, where * denotes the $p$-value is less than 1e-3.}
    \label{tab13}
\end{table*}

\subsection{Empirical Results on FN 1.7}
Table \ref{tab9}, \ref{tab10}, \ref{tab11} list our results with comparing models. \textsc{Kid} outperforms previous state-of-the-art except \citet{su-etal-2021-knowledge}. FN 1.7 is the up-to-date extension version of FN 1.5 containing more fine-grained frames and FEs. However, there are only few models reporting their results on FN 1.7 and we hope that future work on frame semantic parsing can be more focused on FN 1.7.

\subsection{Time Costs of FKG}
The FKG is built over the full FrameNet containing more than 10,000 nodes while the intra-frame and inter-frame relations make the graph larger. Since we need to encode the full FKG when parsing a single sentence, it's necessary to explore the time costs of FKG. Results are shown in Table \ref{tab12} and we can find the time encoding FKG is approximately 20\% in the whole runtime and may slightly hurt the efficiency of our models. However, in inference time, the representations of nodes in FKG are fixed and we can load the representations offline to reduce the inference time.

\subsection{Statistical Analysis of \textsc{Kid} on FN 1.5}
For evaluating solidity of our model, we train \textsc{Kid} with five random seeds. The average performances with deviation and results of significance testing are listed in Table \ref{tab13}. The significance testing is to show whether our model significantly outperforms previous state-of-the-art, and we do not conduct significance testing for \textsc{Kid} (BERT) because we do not outperform \citet{su-etal-2021-knowledge} on frame accuracy. All $p$-values are less than 0.05 and even some $p$-values are less than 1e-3, which proves the solidity of our model.



\end{document}